\documentclass[letterpaper, 10 pt, conference]{ieeeconf}  

\IEEEoverridecommandlockouts 

\usepackage{graphics} 
\usepackage{graphicx}   
\usepackage{subfigure}
\graphicspath{{pics/}} 
\usepackage{booktabs}   
\usepackage{siunitx}
\usepackage{times} 
\usepackage{amsmath} 
\usepackage{xfrac}
\usepackage{mathtools}
\usepackage{amssymb}  
\usepackage{algorithm}
\usepackage{algpseudocode}
\usepackage{bm}
\usepackage{cite}
\usepackage[
    colorlinks=true,
    linkcolor=black,      
    citecolor=black,       
    urlcolor=black,        
]{hyperref}
\usepackage{multirow}
\usepackage{tikz}
\usepackage{pgfplots}
\pgfplotsset{compat=newest}
\usepackage{xcolor}         
\usepackage{lipsum}
\usepackage{cleveref}
\usepackage{pifont}
\usepackage{flushend}

\newcommand{\bmh}[1]{\bm{\hat{#1}}}

\newcommand{\condon}{\,|\,}

\title{\LARGE \bf
From Demonstrations to Safe Deployment: Path-Consistent Safety Filtering for Diffusion Policies
}

\author{
    Ralf Römer$^{\star, 1}$, 
    Julian Balletshofer$^{\star, 1}$, 
    Jakob Thumm$^{2}$,  
    Marco Pavone$^{2}$, \\
    Angela P. Schoellig$^{1}$, and 
    Matthias Althoff$^{1}$%
    \thanks{$^{\star}$Equal contribution.}%
    \thanks{$^{1}$ Department of Computer Engineering, Munich Institute of Robotics and Machine Intelligence (MIRMI), Technical University of Munich, Germany.
    {\tt\small \{ralf.roemer, julian.balletshofer, angela.schoellig, althoff\}@tum.de}}%
    \thanks{$^{2}$ Department of Aeronautics and Astronautics, Stanford University, USA.
    {\tt\small \{thumm, pavone\}@stanford.edu}} %
}

\begin{document}

\maketitle

\begin{abstract}
Diffusion policies (DPs) achieve state-of-the-art performance on complex manipulation tasks by learning from large-scale demonstration datasets, often spanning multiple embodiments and environments.
However, they cannot guarantee safe behavior, requiring external safety mechanisms.
These, however, alter actions in ways unseen during training, causing unpredictable behavior and performance degradation.
To address these problems, we propose path-consistent safety filtering (PACS) for DPs.
Our approach performs path-consistent braking on a trajectory computed from the sequence of generated actions.
In this way, we keep the execution consistent with the training distribution of the policy, maintaining the learned, task-completing behavior.
To enable real-time deployment and handle uncertainties, we verify safety using set-based reachability analysis.
Our experimental evaluation in simulation and on three challenging real-world human-robot interaction tasks shows that PACS (a) provides formal safety guarantees in dynamic environments, (b) preserves task success rates, and (c) outperforms reactive safety approaches, such as control barrier functions, by up to \SI{68}{\percent} in terms of task success. Videos are available at our project website: \href{https://tum-lsy.github.io/pacs/}{tum-lsy.github.io/pacs}.
\end{abstract}

\section{Introduction}
One of the main goals of robotics is to reduce the mental and physical burden on human workers by automating tedious and repetitive tasks. 
Recently, imitation learning using diffusion models~\cite{ho2020denoising} and flow matching~\cite{lipman2022flow} has enabled robots to successfully perform these complex, long-horizon manipulation tasks~\cite{chi2023diffusion, reuss2024multimodal}.
These advances have been reinforced
by the availability of large-scale demonstration datasets~\cite{o2024open} and the emergence of vision-language-action models~\cite{black2024pi_0, shukor2025smolvla}.
Diffusion policies (DPs), i.e., robot policies parameterized by generative diffusion or flow-matching models, can perform tasks in dynamic environments~\cite{ye2025ra, wang2025one}.
However, since DPs are black-box models, they lack safety guarantees for avoiding collisions or safely interacting with dynamic objects.
This prevents their deployment in human-centric environments, where safety has to be formally guaranteed.

\begin{figure}[t]
    \centering
    \vspace{1.7mm}
    \includegraphics[width=0.95\linewidth, trim=0 100 0 10, clip
    ]{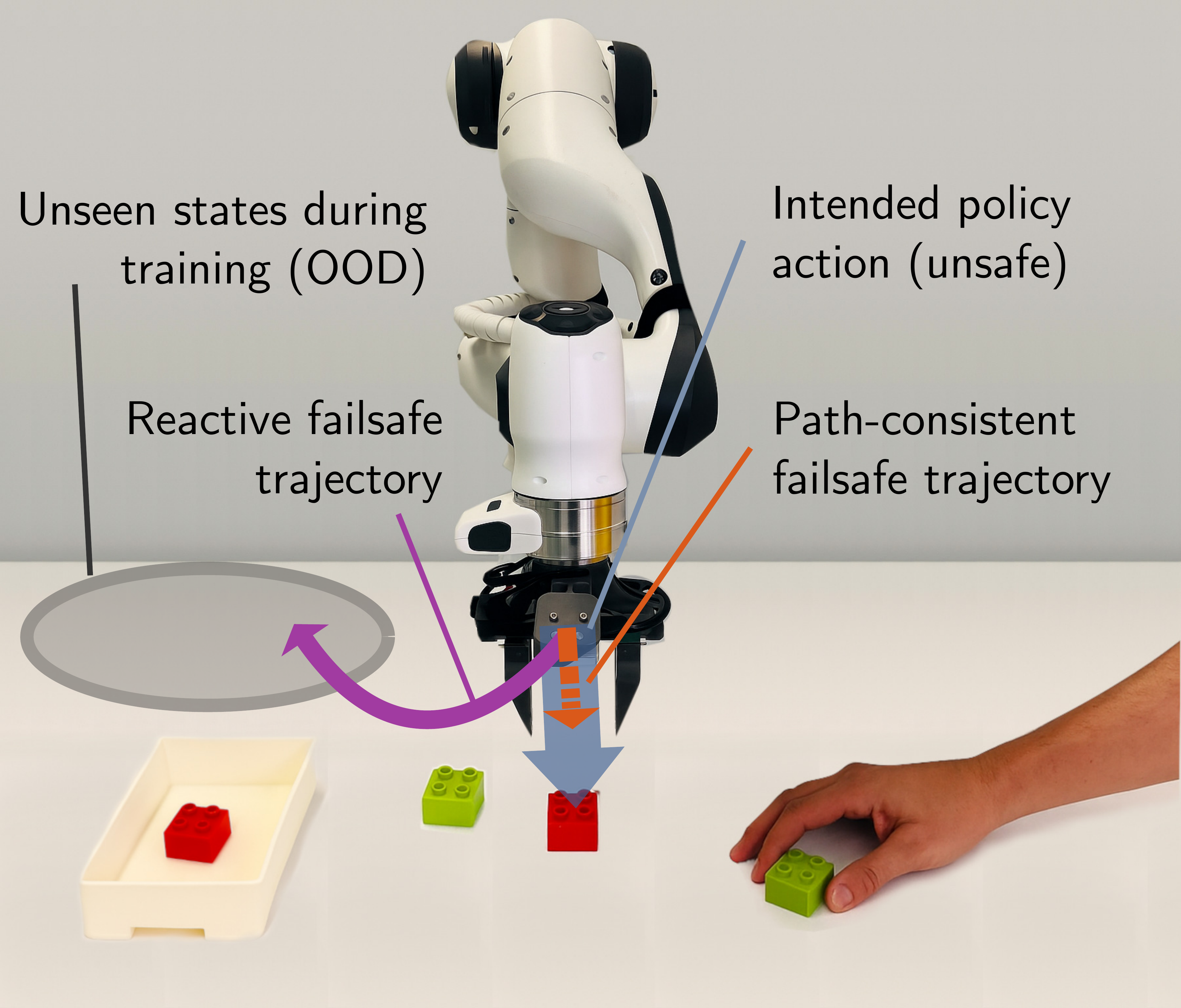}
    \caption{Deploying DPs in dynamic environments with moving objects requires safeguarding mechanisms, as the intended policy actions may be unsafe.
    Reactive strategies, such as control barrier functions, often drive the agent into out-of-distribution (OOD) states not seen during training, leading to unpredictable behavior.
    We propose that safety mechanisms for DPs should remain consistent with the intended path of the robot to avoid out-of-distribution states and preserve high task success rates.
    }
    \label{fig:motivation}
\end{figure}


Many works have shown how safety can be ensured for learning-based policies during deployment~\cite{brunke2022safe, krasowski2023provably}, e.g., using control barrier functions~\cite{ames_2019_ControlBarrier, singletary2022safety} or predictive safety filters~\cite{wabersich_2021_PredictiveSafety, tearle2021predictive}.
Most of these methods are reactive safety mechanisms that satisfy safety constraints by changing the path of the robot away from the dynamic object. 
However, the vast amount of available demonstrations~\cite{o2024open} is not recorded with any of such safety mechanisms in place.
Therefore, as illustrated in~\autoref{fig:motivation}, reactive safety mechanisms often drive the system away from the demonstration data distribution into unseen states.
DPs are particularly prone to errors in these out-of-distribution cases, increasing the risk of task failure~\cite{agia2025unpacking, xu2025can, romer2025failure}.
To avoid this out-of-distribution situation, the interventions of a safety mechanism for DPs should ideally keep the system close to the desired actions of the policy.

We achieve this by proposing a \underline{Pa}th-\underline{c}onsistent \underline{s}afety filter~(PACS)
for DPs.
The core idea is to use path-consistent braking to guarantee safety while keeping the system on its intended path.
To this end, PACS translates the entire action sequence (chunk) generated by the policy into an intended trajectory, along which the robot is slowed down or stopped.
We use set-based reachability analysis~\cite{althoff_2019_Effortless} to verify safety constraints, enabling real-time deployment at~\SI{1}{\kilo\hertz}.

In summary, our main contributions are:
\begin{itemize} 
    \item We perform the first provably safe deployment of DPs for human-robot interaction~(HRI) tasks, representing safety-critical and challenging examples of dynamic real-world environments.
    \item We demonstrate that by avoiding OOD states, PACS achieves significantly higher task success rates than reactive safety filters, both in simulated (\SI{68}{\percent} better) and hardware experiments (\SI{37}{\percent} better).
    \item We demonstrate that an intermediate trajectory generation from action chunks lead to a \SI{28}{\percent} improvement in task success rates compared to sequentially treating each action individually.
\end{itemize}


\section{Related Work}
\paragraph{Safety Filters in Robotics}
Deploying autonomous agents in safety-critical applications requires mechanisms to prevent harmful behavior. 
A common approach is to apply safety filters, which have been extensively reviewed in recent surveys~\cite{brunke2022safe, krasowski2023provably, wabersich_2023_DataDrivenSafety, hsu_2024_SafetyFilter}. 
Safety filters typically verify whether a failsafe action exists that can bring the robot to a safe state after executing the action proposed by the policy~\cite{hsu_2024_SafetyFilter}. 
Several techniques follow this approach: 
Model predictive safety filters~\cite{gros_2020_SafeReinforcement, wabersich_2021_PredictiveSafety} and control barrier functions~\cite{ames_2019_ControlBarrier, marvi_2021_SafeReinforcement, singletary2022safety, morton_2025_SafeTaskconsistent} formulate an optimization problem to compute the safe control input closest to the desired input. 
In contrast, methods based on reachability analysis~\cite{althoff_2015_IntroductionCORA, akametalu_2014_ReachabilitybasedSafe, fisac_2019_GeneralSafety, krasowski_2020_SafeReinforcement, shao_2021_ReachabilitybasedTrajectory, thumm_2022_ProvablySafe, thumm_2025_GeneralSafety} ensure safety by keeping available a provably correct failsafe trajectory. 
These failsafe trajectories can be path-consistent, stopping the robot along its intended path~\cite{zanchettin2013PathconsistentSafety, shao_2021_ReachabilitybasedTrajectory, thumm_2022_ProvablySafe}, or non-path-consistent, guiding the robot away from unsafe regions.

\paragraph{Safety of Diffusion Policies}
Compared to more classical programming-by-demonstration approaches~\cite{zeestraten2016online}, the stochastic and end-to-end nature of diffusion policies makes them particularly prone to unsafe and erratic behavior, especially in unseen situations~\cite{agia2025unpacking, xu2025can, romer2025failure}. 
One way to address this problem is by adding an additional safety layer to the learned policy, e.g., using control barrier functions~\cite{brunke2025semantically}.
Another example is RAIL~\cite{jung2024railreachabilityaidedimitationlearning}, which maintains a failsafe plan based on reachability analysis, defined as a stopping motion that stays as close as possible to the nominal trajectory.
While theoretically well-grounded, such post-processing methods only consider static constraints so far and significantly distort the action distribution, leading to a notable drop in task performance~\cite{jung2024railreachabilityaidedimitationlearning}.
To address this problem, several works directly incorporate safety constraints into the iterative action generation~(denoising) process of diffusion- and flow-based policies.
Adding cost gradients~\cite{carvalho2025motion} or classifier-free guidance~\cite{nikken2024denoising} can steer the generated action sequence towards safe regions, but these methods do not guarantee hard constraint satisfaction.
As an alternative, iterative projections to a safe set~\cite{romer2024safe, romer2025diffusion, liang2025simultaneous} or control barrier functions~\cite{xiao2023safediffuser, huang2025sad} have been injected into the denoising process to provide certain safety guarantees.
However, these approaches still change the policy, and their high computational costs limit their applicability to low-dimensional systems~\cite{romer2024safe, romer2025diffusion} or offline planning~\cite{liang2025simultaneous, xiao2023safediffuser, huang2025sad}.
In summary, real-time safeguarding of DPs in dynamic environments without compromising task success remains an open problem.









\section{Problem Statement}
We aim to ensure a safe deployment of DPs when operating in close proximity to or interacting with dynamic objects, including humans.
More specifically, we consider a robotic system with state~$\bm{x}(t) \in \mathcal{X}$ and input~$\bm{u}(t) \in \mathcal{U}$ at time~$t$ that should perform a given task while adhering to a set of safety constraints.
We assume that the state $\bm{x}$ of the system contains the pose of dynamic objects up to a bounded measurement error and that the objects have a bounded input set, e.g., a maximal velocity.
We assume the availability of a set of trajectories demonstrated by experts~${\mathcal{D} = \{\bm{o}_k,\bm{a}_k\}_{k=0}^{K}}$ in the form of {observation-action} pairs. 
The observations $\bm{o} \in \mathcal{O}$ may include robot joint angles, camera images, language commands, and other environmental information.
The action~${\bm{a} \in \mathcal{A}}$ is a high-level goal that the robot should reach, e.g., a desired joint configuration.
From the expert demonstrations, a DP~$\pi: \mathcal{A}^H \times \mathcal{O} \rightarrow \mathbb{R}_{+}$ is trained that returns a distribution over action chunks~$\bm{A}\in\mathcal{A}^H$ of length~$H$.
In this work, we aim to design a controller ${\bm{k}: \mathcal{X} \times \mathcal{A}^H \times \mathbb{R} \rightarrow \mathcal{U}}$ that takes as input the latest action chunk~$\bm{A}$ generated by the policy and returns a safe control input~$\bm{u}_\text{safe}(t)$.


\section{Preliminaries}

\begin{figure*}[t!]
    \centering
    \vspace{1.7mm}
    \includegraphics[width=0.96\linewidth]{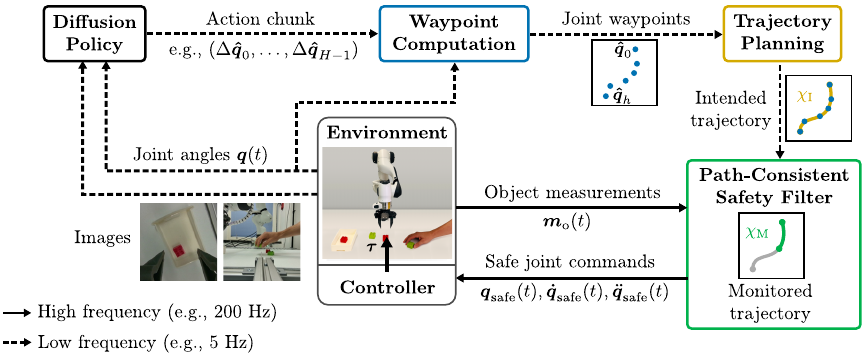}
    \caption{System overview of our proposed path-consistent safety filter (PACS). The policy, conditioned on visual observations and proprioceptive inputs, generates action chunks that are transformed into a sequence of desired waypoints. From these waypoints, we compute a kinematically and dynamically feasible intended trajectory. PACS continuously monitors this trajectory and applies high-frequency safety filtering using reachability analysis to enforce task-specific safety constraints (e.g., collision avoidance or impact force limits). 
    }
    \label{fig:overview_method}
\end{figure*}
This section describes the technical background on DPs and reachability-based safety verification. Throughout the paper, we use~$(\cdot)$ as a placeholder for the arguments of a probability distribution or function.

\subsection{Imitation Learning with Diffusion Policies}
The demonstration data~$\mathcal{D}$ encodes the desired behavior for solving the task.
A DP~$\pi$ is trained to approximate the conditional action distribution in the demonstrations~\cite{chi2023diffusion}.
Once trained, DPs and VLAs generate new actions by iteratively transforming a random noise sample into the learned target distribution~\cite{ho2020denoising, lipman2022flow}.
Most DPs return a chunk of~$H$ consecutive actions~${\bm{A}_k = (\bm{a}_{k},\bm{a}_{k+1},\dots,\bm{a}_{k+H-1})}$ instead of only the next action to improve the temporal consistency and execution speed~\cite{ chi2023diffusion, zhao2023learning}.
During deployment, we can iteratively sample new action chunks~${\bm{A}_k \sim \pi(\cdot\condon\bm{o}_k)}$, apply the first~$h \leq H$ actions~$\bm{a}_{k:k+h-1}$ to the system for a time~$\Delta t$, respectively, and replan after a time~$h\Delta t$ has passed.
This general formulation of imitation learning policies encompasses state-of-the-art generative single-task policies~\cite{chi2023diffusion, reuss2024multimodal} and generalist vision-language-action models~\cite{black2024pi_0, shukor2025smolvla}.

\subsection{Safety Constraints for Dynamic Objects}\label{sec:safety_constraints}
The robot and the objects are considered as dynamic systems, and we define the trajectory of a system at time~$t$ for a given initial state~${\bm{x}_0 \in \mathcal{X}_0}$ and a possible input trajectory~${\bm{u}(\cdot)}$ by ${\chi(t;\bm{x}_0;\bm{u}(\cdot))}$. Following~\cite[p.~5]{althoff_2019_Effortless}, we formally introduce the reachable set~${\mathcal{R}(t)}$ of a system at time~$t$ for a compact set of possible inputs~$\mathcal{U}$ by
\begin{equation}
    \mathcal{R}(t) \coloneq \{\chi(t;\bm{x}_0;\bm{u}(\cdot)) \condon \bm{x}_0 \in \mathcal{X}_0, \forall t : \bm{u}(t)\in\mathcal{U} \},
\end{equation}
and denote all states that are reachable during a time interval as~${\mathcal{R}([t_a, t_b]) = \bigcup_{t \in [t_a, t_b]} \mathcal{R}(t)}$.
We denote the \textit{reachable occupancy} by $\mathcal{C}(t)$. For the reachable occupancy computation of the robot $\mathcal{C}_\text{robot}(t)$, we assume a bounded tracking error of the desired trajectory. This assumption could be replaced by applying more advanced methods as presented in \cite{stefan2025ControllerSynthesis}.
To compute the reachable occupancies of dynamic objects $\mathcal{C}_\text{object}(t)$, we assume access to measurements of positions of objects with bounded measurement errors. Moreover, we consider velocity and acceleration limits, and further include sensor delay and bounded measurement errors on position and velocity, following the approach in \cite{althoff_2019_Effortless}.

Following prior work~\cite{althoff_2019_Effortless, liu_2021_OnlineVerification, beckert_2017_multiple_safety}, we define a collision between an object and the robot if their reachable occupancies $\mathcal{C}_{\text{object}}(t)$ and $\mathcal{C}_{\text{robot}}(t)$ intersect, i.e., a collision event is defined as
\begin{equation}
\label{eq:collision}
    c_{\text{coll}}(t) \coloneq \mathcal{C}_{\text{object}}(t) \cap \mathcal{C}_{\text{robot}}(t) \not= \emptyset.
\end{equation}
Depending on the task and object, collisions should either be avoided altogether or restricted to non-harmful, low-force interactions.
Accordingly, we define two types of safety constraints 
in accordance with ISO/TS 15066:2016.
Power and force limiting (PFL) targets human-robot collaboration with close interaction and allows contacts with humans as long as the kinetic energy of the robot is below pain and injury thresholds~$T_{\text{safe}}(t)$, resulting in the PFL safety constraint 
\begin{align}
\label{eq:PFL-constraint}
    c_{\text{safe}, \text{PFL}}(t) &\coloneqq  \overline{c_{\text{coll}}(t)} \, \lor \big(T_{\text{robot}}(t) \le T_{\text{safe}}(t)\big).
\end{align}
The safe energy threshold~$T_{\text{safe}}(t)$ can depend on the collision type (constrained collisions for clamping or unconstrained collisions for free movement), the shape of the robot part in contact (e.g., sharp or blunt edges), or the human body part affected in the worst case~\cite{kirschner_2024_constrained,kirschner_2024_unconstrained, thumm_2025_GeneralSafety}. 
In contrast, speed and separation monitoring (SSM) does not permit any contact with humans, targeting coexistence without close interaction.
If a collision can occur, the robot must be at a complete stop with zero kinetic energy, i.e., $T_{\text{safe}}(t)=0$ in~\eqref{eq:PFL-constraint}.


\section{Path-Consistent Safety Filter}
We now detail PACS, which ensures the safe deployment of pretrained DPs. An overview is provided in~\autoref{fig:overview_method}.

\subsection{Safety-Permitting Diffusion Policy}
To maintain high task performance despite path-consistent safety interventions, we need to design the policy in such a way that slowing down or stopping the robot does not in itself represent an out-of-distribution situation.
To achieve this, we only include RGB images~$\bm{I}(t)$ and joint angles~$\bm{q}(t)$ into the observation~${\bm{o}(t)}$ and no explicit velocity information, which is a common approach in recent works~\cite{black2024pi_0, shukor2025smolvla}.
Our approach is applicable to various action spaces, but we restrict our presentation to joint velocity control for clarity.
In this work, the actions are the delta joint positions~${\bm{a}_i = \Delta \bmh{q}_i = (\bmh{q}_{i+1}-\bmh{q}_{i})}$
and we use~$\hat{\cdot}$ to distinguish the desired actions generated by the policy from the actual safe commands sent to the robot. Without loss of generality, we reset the clock in each step of the planning algorithm and safety filter, i.e., $t_0=0$.
By integrating the desired joint positions forward in time, starting from the current joint position $\bm{q}(0)$,
we compute a sequence of waypoints
\begin{align}
    \label{eq:policy_desired_path}
    \bmh{p}(0) = (\bmh{q}_0,\dots,\bmh{q}_h), \quad \text{where} \quad \bmh{q}_0 = \bm{q}(0),
\end{align}
describing the desired path (Algorithm~\ref{alg:shielded-diffusion}, line~6).

{
\begin{algorithm}[t!]
\caption{Path-consistent safety filter for diffusion policies~(DPs) and vision-language-action models~(VLAs).}
\label{alg:shielded-diffusion}
\begin{algorithmic}[1]
\Require DP or VLA~$\pi(\cdot)$, chunk length~$H$, safety filter horizon $h$, policy timestep $\Delta t$, safety step size $\alpha_{s}$, initial configuration $\bm{q}(0)\in \mathcal{X}_{\mathrm{Safe}}$
\State $t \gets 0$
\State $\textit{success} \gets \textbf{False}$
\While{$\neg \textit{success}$} \Comment{Policy steps}
    \State $\bm{o}(t) \gets \Call{GetObservation}{t}$
    \State $\bm{A}(t) = (\Delta \bmh{q}_0,\ldots,\Delta \bmh{q}_{H-1}) \sim \pi(\cdot\condon\bm{o}(t))$
    \State $(\bmh{q}_0,\dots,\bmh{q}_h) \gets \Call{Integrate}{\bm{A}(t)}$
    \State $\chi_I([t_0,t_h]) \gets \Call{ComputeTrajectory}{(\bmh{q}_0,\ldots,\bmh{q}_h)}$
    \State $\textit{shield} \gets \Call{SetTrajectory}{\chi_I([t_0,t_h])}$
    \For{$k \gets 0$ \textbf{to} $\frac{h\Delta t}{\alpha_s}$} \Comment{Safety steps}
        \State $\bm{m}_{\mathrm{o}}(t) \gets \Call{GetObjectMeasurements}{t}$
        \State $\bm{u}_{\mathrm{safe}}(t + \alpha_s) \gets \Call{ShieldStep}{t,\,\bm{m}_{\mathrm{o}}(t)}$
        \State $\Call{ApplyAction}{\bm{u}_{\mathrm{safe}}(t + \alpha_s)}$
        \State $t \gets t + \alpha_s$
    \EndFor
    \State $\textit{success} \gets \Call{CheckSuccess}{\bm{o}(t)}$
\EndWhile
\end{algorithmic}
\end{algorithm}}

\subsection{Path-Consistent Safety for Diffusion Policies}
To ensure that our safety filter does not steer the policy into out-of-distribution states, we adopt a path-consistent safety formulation.
The key idea is to guarantee safety by modifying the joint speed~$\dot{\bm{q}}(t)$ and acceleration~$\ddot{\bm{q}}(t)$ while staying on the intended trajectory. 
Existing path-consistent safety filters for learning-based policies using reachability analysis~\cite{thumm_2025_GeneralSafety, thumm_2022_ProvablySafe} can only take as action input a single goal position with zero velocity. 
However, DPs and vision-language-action models predict a chunk~$\bm{A}(t)$ of multiple consecutive actions together and operate at a relatively low update frequency~$1/(h\Delta t)$.
Therefore, directly applying single-action approaches~\cite{thumm_2025_GeneralSafety, thumm_2022_ProvablySafe} to these policies either leads to a very slow execution by stopping at each waypoint or leaves out intermediate action steps, sacrificing fine-grained control and causing a deviation from the desired behavior.

To construct smooth trajectories that align with the predicted action chunk, we propose an intended trajectory planning module~(cf.~\autoref{fig:overview_method}).
For each action chunk~$\bm{A}(t)$ starting from the initial time step $t_0=0$, we compute a time-optimized intended trajectory~${\chi_I([0, t_h])=\big\{\bm{x}(\tau) \, \big| \, \tau \in [0, t_h]\}}$ (Algorithm~\ref{alg:shielded-diffusion}, line 7), where~$\bm{x}(\tau) = \big(\bm{q}(\tau), \dot{\bm{q}}(\tau), \ddot{\bm{q}}(\tau), \dddot{\bm{q}}(\tau)\big)$ is the state of the system, by solving the optimization problem
%
%
\begin{subequations}
\label{eq:shield_intended_trajectory}
\begin{align}
\label{shield:path_consistency_condition}
\min_{t_h,\, \chi_I([0, t_h])} \quad & t_h\\
\text{s.t.} \qquad 
& \forall \tau \in [0, t_h]:\; \bm{x}(\tau) \in \mathcal{X}, \\
\label{shield:path_kinemtic_condition}
& \forall k \in \{0,\dots,h\}: \bm{q}(t_k) = \bmh{q}_k,
\end{align}
\end{subequations}
where the set~$\mathcal{X}$ defines robot-specific kinematic and dynamic constraints 
$\bm{q}_{\min} \le \bm{q}(\tau) \le \bm{q}_{\max}$, $\|\dot{\bm{q}}(\tau)\| \le \dot{q}_{\max}$, $ 
\|\ddot{\bm{q}}(\tau)\| \le \ddot{q}_{\max}$, and $\|\dddot{\bm{q}}(\tau)\| \le \dddot{q}_{\max}$. 
Adjusting the velocity, acceleration, and jerk limits allows us to adapt the execution speed of the action chunk, which gives us a more fine-grained control over the execution speed of the policy compared to modifying the actions directly.
Importantly, our intermediate planning module provides a trajectory by solving~\eqref{eq:shield_intended_trajectory}, which we forward to our path-consistent safety filter (Algorithm~\ref{alg:shielded-diffusion}, line 8).

The safety filter (Algorithm~\ref{alg:shielded-diffusion}, line 9-14) operates at a high frequency and formally guarantees constraint satisfaction while staying on the desired path~\eqref{eq:policy_desired_path}.
Based on the current state of the robot~$\bm{x}_0 = \bm{x}(0)$ and the measured object positions~$\bm{m}_{\text{o}}(t)$ (Algorithm~\ref{alg:shielded-diffusion}, line 10), we compute a path-consistent monitored trajectory \cite{thumm_2025_GeneralSafety} by concatenating an intended trajectory (desired or recovery motion) $[0,t_\text{I}]$ and a failsafe trajectory (stopping motion) $[t_\text{I},t_\text{F}]$
\begin{equation}
\label{eq:shield_monitored_trajectory}
\chi_\text{M}(t) =
\begin{cases}
\chi_I\big(\tau;\, \bm{x}_0, \bm{u}_I(\cdot)\big), & \tau \in [0, t_\text{I}], \\[0.4em]
\chi_F\big(\tau;\, \bm{x}_I,\bm{u}_F(\cdot)\big),&\tau\in[t_\text{I}, t_\text{F}].
\end{cases}
\end{equation}
At each safety timestep of duration~$\alpha_\text{s}$, we verify the monitored trajectory~\eqref{eq:shield_monitored_trajectory} for the task-specific safety constraints as described in~\Cref{sec:safety_constraints}.
If the verification is successful, we execute the intended trajectory~${\bm{u}_\text{safe}(t)=\bm{u}_\text{I}(t)}$; otherwise, we proceed with the last successfully verified failsafe trajectory that brings the system to an invariable safe set, i.e., ${\bm{u}_\text{safe}(t)=\bm{u}_\text{F}(t)}$.
We repeat this process until receiving a new action chunk from the policy after~$h\Delta t / \alpha_\text{s}$ safety steps.
Since this safeguarding mechanism is provably safe by induction, as shown in prior work~\cite{thumm_2025_GeneralSafety}, PACS enables the safe deployment of DPs. 

\begin{figure*}[ht]
\vspace{1.7mm}
    \centering
    \subfigure[\textsc{Sorting:} Put the red blocks into the box while a person is taking the green blocks.]{%
        \centering
        \includegraphics[height=0.3\textwidth, trim={0cm 8cm 0cm 4cm}, clip]{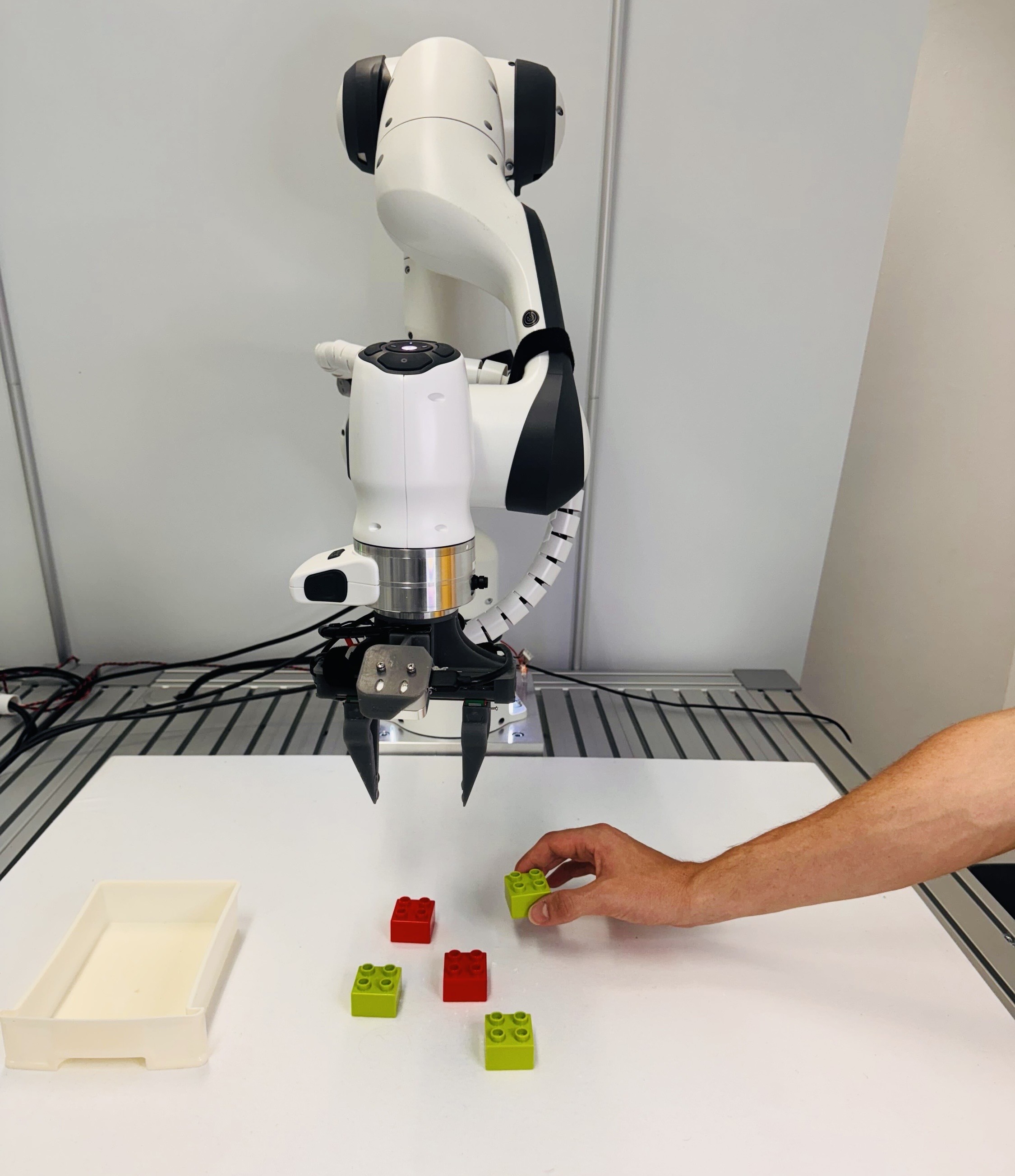}%
        \label{fig:sorting}%
    }\hfill
    \subfigure[\textsc{Handover:} Grab blocks from a moving hand and put them into the box.]{%
        \centering
        \includegraphics[height=0.3\textwidth, trim={0cm 10cm 0cm 2cm}, clip]{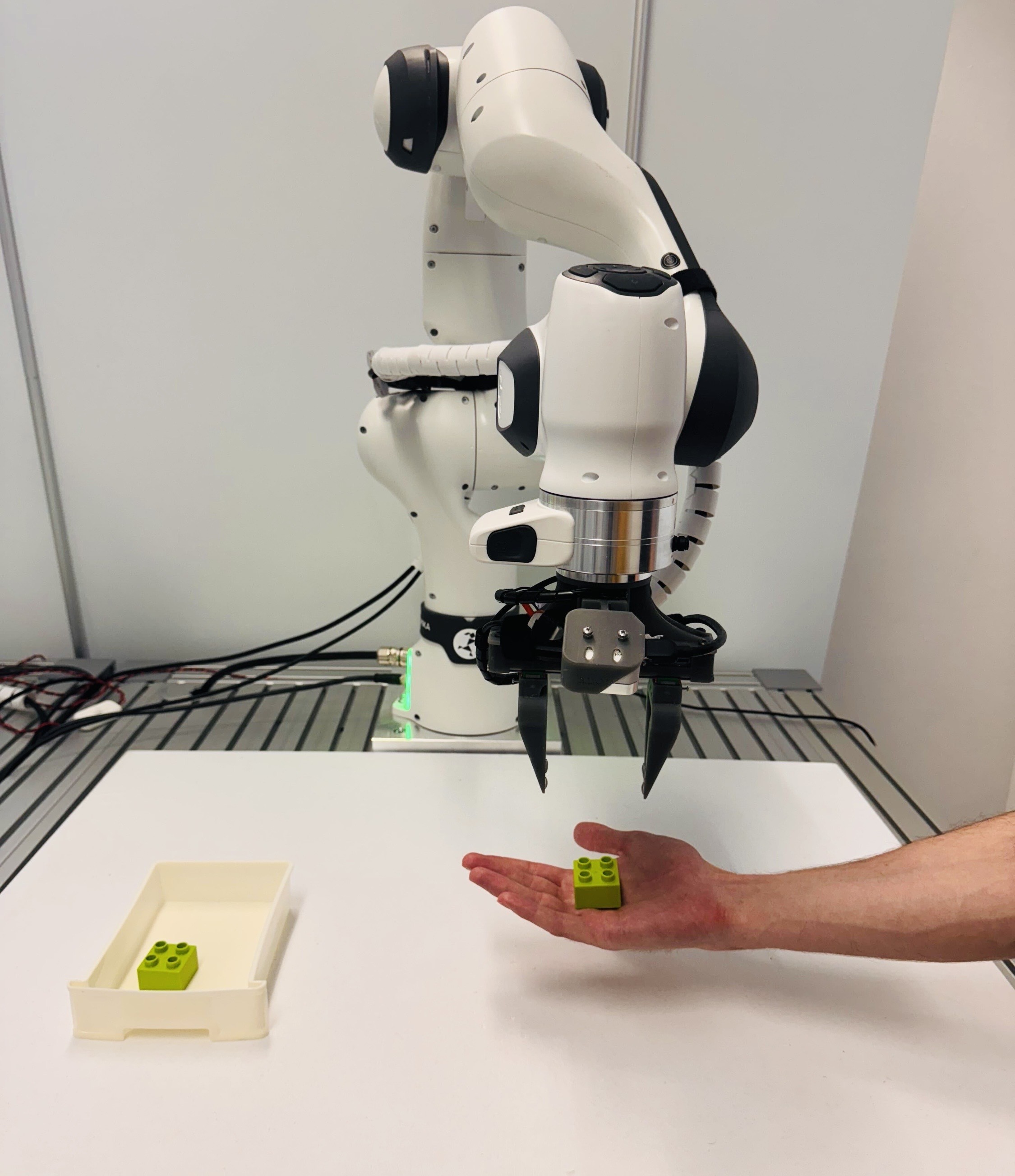}%
        \label{fig:handover}%
    }\hfill
    \subfigure[\textsc{Feeding:} Carefully put a fork with food in the mouth and pull the fork back out.]{%
        \centering
        \includegraphics[height=0.3\textwidth, trim={0cm 0cm 0cm 13cm}, clip]{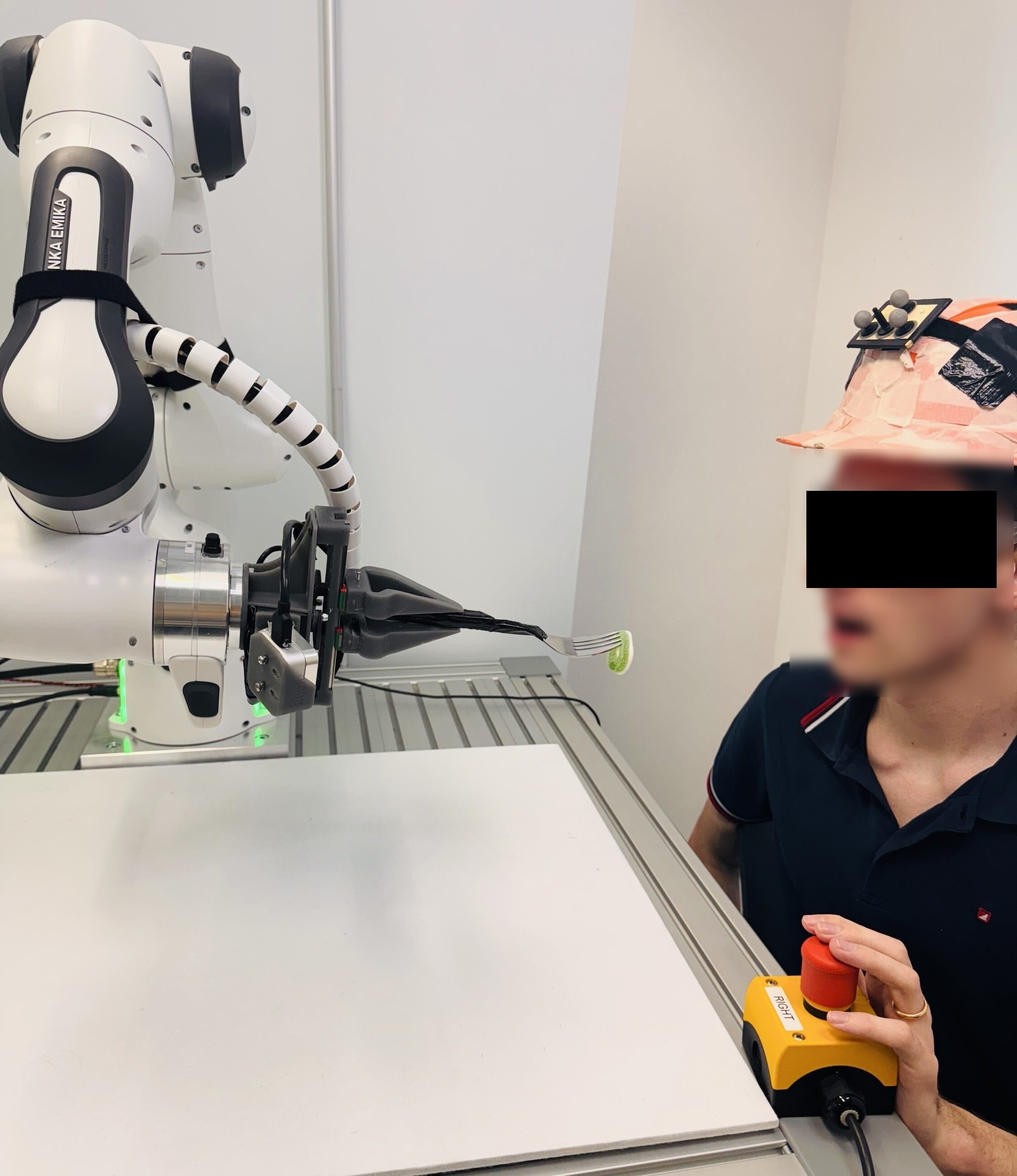}%
        \label{fig:feeding}%
    }
    \caption{Visualization of our three real-world tasks, which require safe and reactive motion in close proximity to the human body. \textsc{Sorting} represents a coexistence task, where no collisions are permitted, whereas~\textsc{Handover} and~\textsc{Feeding} are collaboration tasks requiring non-harmful, low-force interactions.}
    \label{fig:tasks}
\end{figure*}

\section{Evaluation}
We conduct extensive experiments in simulation and on hardware to evaluate the real-time safeguarding capabilities of our method.
Our code will be released upon acceptance to preserve anonymity.
Our quantitative evaluation primarily aims to validate the following hypotheses:
\begin{itemize}
    \item[\textbf{H1}] Through behavior cloning from demonstrations alone, DPs cannot learn a safe behavior for interaction with moving objects. 
    \item[\textbf{H2}] With our path-consistent safeguarding method, the task success rate remains close to the unsafe policy ($\pm$ \SI{5}\%) in our conducted experiments.
    \item[\textbf{H3}] Computing intended trajectories from entire action chunks leads to higher task success rates than single-action approaches.
    \item[\textbf{H4}] PACS is real-time capable and does not reduce task execution speed, unless required to ensure safety. 
\end{itemize}

\subsection{Simulation Experiments}
{
\setlength{\tabcolsep}{5pt}
\begin{table}[tb!]
\centering
\caption{Robomimic task success rate results on \num{100} rollouts.}
\label{tab:method_comparison}
\begin{tabular}{lccccc}
\toprule
Method & Safe? & \textsc{Lift} & \textsc{Can} & \textsc{Square} & Average \\
\midrule
Operational space contr. & \ding{53} & \textbf{1.00} & \textbf{0.99} & \textbf{0.74} & \textbf{0.91} \\
OFF & \ding{53} & 0.92 & 0.83 & 0.34 & 0.70 \\
\midrule
Contr. barrier func.~\cite{singletary2022safety} & \ding{51} & 0.11 & 0.00 & 0.00 & 0.04 \\
Single-action - SSM & \ding{51} & \textbf{0.97} & 0.26 & 0.00 & 0.41 \\
Single-action - PFL & \ding{51} & 0.94 & 0.33 & 0.04 & 0.44 \\
PACS \textbf{(ours)} - SSM & \ding{51}  & \textbf{0.97} & 0.80 & 0.30 & 0.69 \\
PACS \textbf{(ours)} - PFL & \ding{51}  & 0.93 & \textbf{0.85} & \textbf{0.38} & \textbf{0.72} \\
\bottomrule
\end{tabular}
\end{table}}

First, we evaluate the performance of PACS against an unsafe operational space controller and a control barrier function baseline~\cite{singletary2022safety} on the three robomimic~\cite{mandlekar2021matters} benchmark tasks \textsc{Lift}, \textsc{Can}, and \textsc{Square}.
We integrated these tasks in human-robot-gym~\cite{thumm_2024_human_robot_gym} and added a spherical dynamic object that moves in pre-defined patterns on the surface of the table. 
The goal is to fulfill the original task while ensuring the safety constraints $c_{\text{safe}, \text{SSM}}$ (for operational space controller, control barrier function, and SSM) or $c_{\text{safe}, \text{PFL}}$ (for PFL).
To evaluate~\textbf{H3}, we further tested a single-action version of PACS that sequentially computes new intended trajectories from each action in the action chunk.

We trained the DP with a temporal UNet architecture~\cite{chi2023diffusion} on the provided dataset of~\num{200}~proficient human teleoperated demonstrations per task.
The policy receives the end-effector and object poses of the last two timesteps as observations and predicts desired changes in the end effector pose and gripper opening as actions.
The operational space controller directly controls the robot in Cartesian space to follow these goals.
For all other approaches, we used an inverse kinematics solver~\cite{coumans_2016_PyBulletPython} to translate the desired changes in end-effector pose into desired joint state actions and control the robot in joint space.
Additional implementation details can be found in~\autoref{tab:policy_parameters}.
For the operational space controller baseline, we used $h = 8$ action execution steps and applied each action for $\Delta t = \SI{50}{\milli \second}$.
To evaluate the impact of the safety mechanism and the inverse kinematics translation separately, we also tested the performance of our safety shield in OFF mode.
Here, we compute the same intended trajectory in joint space, but never execute a failsafe trajectory. 

The results in~\autoref{tab:method_comparison} highlight that PACS achieves a significantly higher task success rate than the reactive control barrier function safety filter. 
Comparing the safe SSM and PFL mode with the controller in the OFF setting, we see that path-consistent braking has no significant impact on the performance of the policy, \textbf{confirming H2}.
However, if we compare OFF to the operational space controller, we see a noticeable drop in performance in the \textsc{Square} task, which is most likely due to slight inaccuracies induced by transforming the end-effector poses into joint space.
\autoref{tab:method_comparison} further shows that the action chunk trajectories of PACS improve the performance of the sequentially executed single actions by \SI{28}{\percent}, \textbf{confirming H3}. 

{\setlength{\tabcolsep}{4.5pt}
\begin{table}[tb!]
\centering
\caption{Constraints of our real-world tasks.}
\label{tab:energy-thresholds}
\begin{tabular}{@{}lcccc@{}}
\toprule
Task & Constraint Type & Body Part & Condition & $T_\text{safe}$ [\si{J}] \\ 
\midrule
\textsc{Sorting} & SSM & Hand & - & 0.000 \\
\midrule
\multirow{2}{*}{\textsc{Handover}} 
 & \multirow{2}{*}{PFL} & \multirow{2}{*}{Hand} & Constrained & 0.014 \\
 & &                      & Unconstrained & 0.265 \\
\midrule
\textsc{Feeding} & PFL & Head (Eye) & Unconstrained & 0.001 \\
\bottomrule
\end{tabular}
\end{table}}

\subsection{Real-World Setup}
Second, we conducted hardware experiments with a Franka FR3 manipulator for HRI, as safety is particularly critical in this area. 
\subsubsection{Tasks}
We considered three tasks requiring safe motion in close proximity to humans, as visualized in~\autoref{fig:tasks}: 
\begin{itemize}
    \item \textsc{Sorting:} We placed two red and three green blocks randomly on the table. The robot must put the red blocks into a box while the human is taking the green blocks. 
    We investigated different human motion patterns and speeds.
    To ensure safety, no collisions between the robot and the human are allowed ($c_{\text{safe}, \text{SSM}}$).
    \item \textsc{Handover:} The robot must pick a block from the hand of a human partner. To ensure safety, the energy of the impact between the gripper and the hand must not exceed a given energy limit ($T_{\text{safe}, \text{hand}}$). 
    \item \textsc{Feeding:} The robot must put a fork with food into the mouth of the user. 
    To ensure safety, the energy of the impact between the head and the fork must not exceed a given limit ($T_{\text{safe}, \text{head}}$). Note that \mbox{$T_{\text{safe}, \text{head}} \ll T_{\text{safe}, \text{hand}}$}.
    We use a printed AI-generated face with a cut-out mouth for our quantitative experiments and conduct qualitative tests with a real human.
\end{itemize} 

\subsubsection{Safety Constraints}
We used the SaRA framework~\cite{sven_2022_SaRA} to compute reachable occupancies and performed real-time intersection checking.
The safe energy thresholds in~\eqref{eq:PFL-constraint} have been formally identified for the different tasks~\cite{thumm_2025_GeneralSafety}.
Further details and numerical values are provided in~\autoref{tab:energy-thresholds}.
We leveraged Ruckig~\cite{Ruckig} to compute the intended trajectory~\eqref{eq:shield_intended_trajectory}, using the standard joint limits of the FR3, velocity limits between \SI{1.25}{} and \SI{2}{\radian \per \second}, an acceleration limit of \SI{10}{\radian \per \second \squared} and a jerk limit of \SI{400}{\radian \per \second \cubed} for each joint.


{
\setlength{\tabcolsep}{4pt}
\begin{table}[t]
    \vspace{1.7mm}
    \centering
    \caption{Implementation details for the DP and vision-language-action model in our experiments.}
    \begin{tabular}{@{}lccc@{}}
    \toprule 
        Hyperparameter & DP (sim) & DP (real) & SmolVLA \\ 
        \midrule
        Demonstrat. per task & 200 & 50 & 50 \\
        Generative model & Diffusion & Diffusion & Flow Matching \\
        Architecture & U-Net & U-Net & VLM + Transformer \\
        Training steps & 64k 
        & 20k & 40k \\
        Learning rate & ${\text{1} \times \text{10}^{-\text{4}}}$ & ${\text{1} \times \text{10}^{-\text{4}}}$ & ${\text{1} \times \text{10}^{-\text{4}}}$ \\
        Batch size & 100 & 64 & 64 \\
        Action chunk len.~$H$ & 16 & 16 & 50 \\
        Action exec. steps~$h$ & 2 & 6 & 6 \\
        Inference timesteps & 100 (DDPM) & 10 (DDIM) & 10 (Euler) \\
        Sampling time~$\Delta t$ & 100 \si{\milli \second} & 33 \si{\milli \second} & 33 \si{\milli \second} \\
        \bottomrule
    \end{tabular}
    \label{tab:policy_parameters}
\end{table}}
\subsubsection{Policies}
We deployed two generative policies: DP~\cite{chi2023diffusion} using denoising diffusion~\cite{ho2020denoising} and SmolVLA~\cite{shukor2025smolvla} using flow matching~\cite{lipman2022flow}.
The observations consisted of joint angles and the images of a wrist and a third-person camera, and the actions were desired changes in joint angles and the gripper opening.
We provide details on the training data and the policy implementation in~\autoref{tab:policy_parameters}.
CRISP~\cite{pro2025crispcompliantros2} has been used for data recording and for translating the joint commands to motor torques.
The SmolVLA policy struggled with \textsc{Sorting} and \textsc{Feeding}, likely due to the high precision required; therefore, we only evaluated it on the~\textsc{Handover} task.

\subsubsection{Metrics}
We report the success rate of completing the task~(Success) and the success rate of completing the task while remaining safe at all times (Safe Success) based on \num{30} test rollouts.
Moreover, we report the proportion of timesteps for which the task-specific safety constraints are violated (Safety Viol.) and the duration of successful rollouts.

\subsection{Real-World Results}
We deployed the policies with our safety filter OFF and active, similar to the simulation experiments. 
As shown by the results in~\autoref{tab:safety_comparison}, the policies achieved very similar task success rates of about~\SI{80}{\%} on average with and without our safety filter, \textbf{supporting~H2}.
The safeguarded policies do not exhibit constraint violations in our experiments.
In contrast, the nominal policies are in an unsafe state \SI{56}{\%} of the time with each rollout containing constraint violations, resulting in a Safe Success of \SI{0}{\%}, which \textbf{confirms~H1}.
These results highlight the necessity of online safety mechanisms when deploying DPs in dynamic environments and the effectiveness of PACS for this purpose.
{\setlength{\tabcolsep}{2.9pt}
\begin{table}[tb!]
\vspace{1.7mm}
\centering
\caption{Impact of safeguarding in our real-world experiments.}
\begin{tabular}{@{}lllccc@{}}
    \toprule 
    Task & Policy & Safety Filter & Success & Safe Succ. & Safety Viol. \\ 
    \midrule
    \multirow{2}{*}{\textsc{Sorting}} 
        & \multirow{2}{*}{DP} & OFF     & 0.77 & 0.00 & 0.67 $\pm$ 0.12 \\ 
        &                     & PACS (\textbf{ours}) & 0.80 & 0.80 & 0.00 $\pm$ 0.00 \\ 
    \midrule
    \multirow{4}{*}[-0.2em]{\textsc{Handover}} 
        & \multirow{2}{*}{DP} & OFF     & 1.00 & 0.00 & 0.32 $\pm$ 0.07 \\     
        &                     & PACS (\textbf{ours}) & 0.97 & 0.97 & 0.00 $\pm$ 0.00 \\ 
        \cmidrule{2-6}
        & \multirow{2}{*}{SmolVLA} 
                              & OFF     & 0.77 & 0.00 & 0.41 $\pm$ 0.15 \\
        &                     & PACS (\textbf{ours}) & 0.80 & 0.80 & 0.00 $\pm$ 0.00 \\ 
    \midrule
    \multirow{2}{*}{\textsc{Feeding}} 
        & \multirow{2}{*}{DP} & OFF     & 0.63 & 0.00 & 0.85 $\bm{\pm}$ 0.04 \\     
        &                     & PACS (\textbf{ours}) & 0.63 & 0.63 & 0.00 $\pm$ 0.00 \\ 
    \midrule
    \multicolumn{2}{c}{\multirow{2}{*}{Average}}
        & OFF & \textbf{0.79} & 0.00 & 0.56 $\bm{\pm}$ 0.21 \\     
        & & PACS (\textbf{ours}) & \textbf{0.80} & \textbf{0.80} & \textbf{0.00 $\bm{\pm}$ 0.00} \\  
    \bottomrule
\end{tabular}
\label{tab:safety_comparison}
\end{table}}

\begin{figure}
    \centering
    \vspace{1.7mm}
    \begin{tikzpicture}
        \node[inner sep=0pt] (img) {\includegraphics[width=\linewidth]{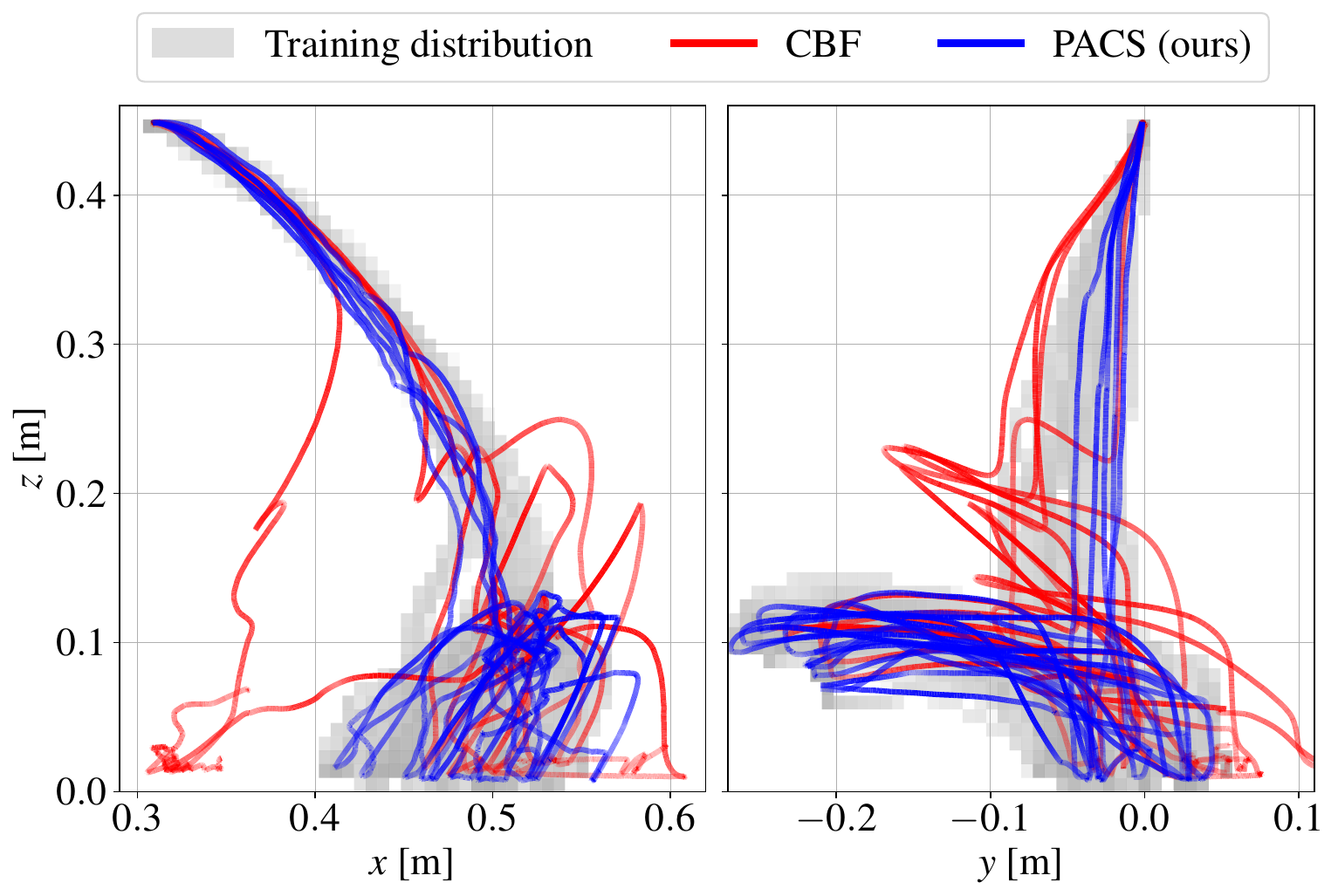}};

        \node[anchor=south] at (-2.95,-0.1) {\shortstack{OOD\\Failures}};
        \draw (-2.95,-0.1) -- (-3,-1.1);
        \draw[fill=yellow, fill opacity=0.4, draw=none] (-3,-1.7) ellipse [x radius=0.5cm, y radius=0.5cm];
        \draw[fill=yellow, fill opacity=0.4, draw=none] (3.9,-1.7) ellipse [x radius=0.25cm, y radius=0.5cm];
    \end{tikzpicture}
    \caption{End-effector paths for the \textsc{Sorting} task. The color intensity of the trajectories indicates the velocity, and the shaded grey areas visualize the training distribution.
    Our safety filter reduces the velocity without leaving the desired path when the human is nearby. In contrast, the control barrier function~(CBF) pushes the robot away from unsafe states, which often leads to out-of-distribution~(OOD) states from which the policy cannot recover.}
    \label{fig:path_comparisons}
\end{figure}
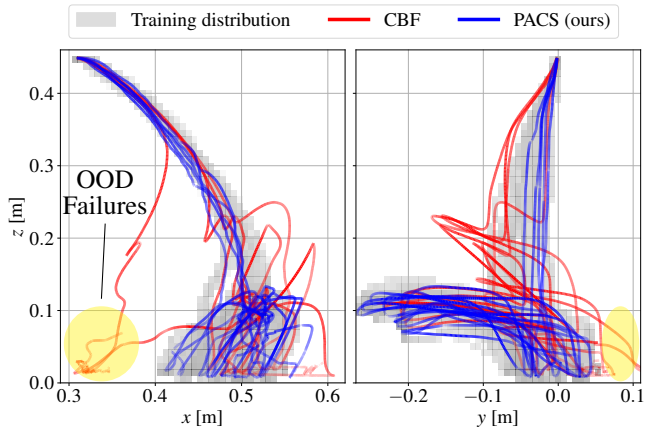

We also compare our path-consistent safety approach with a popular reactive method: control barrier functions~\cite{singletary2022safety}. 
Since control barrier functions are better suited for coexistence than for collaboration tasks, we conduct this comparison for the \textsc{Sorting} task.
As can be seen from~\autoref{tab:sorting_results}, PACS achieved a \SI{37}{\%} higher success rate than the control barrier function-based filter. 
To further examine this inferior performance, we visualize the training data distribution and the end-effector positions of randomly selected rollouts  in~\autoref{fig:path_comparisons}.
While our method kept the policy in distribution, the control barrier function often pushed the robot into out-of-distribution states from which the policy struggles to recover, leading to a higher failure rate.
Based on our simulation results, we find that PACS overall achieved a \SI{51}{\%} higher task success rate than the safety filter using a control barrier function.
Besides, we observe that policy recovery usually takes a few seconds, leading to a higher average task completion time.
The average computation time of one safety step with PACS was~\SI{0.20}{\milli\second} compared to~\SI{0.64}{\milli\second} for the control barrier function, and recalculating the intended trajectory took about~\SI{5}{\milli \second}.
This \textbf{supports~H4} and shows that our method can be deployed in real time in highly dynamic environments.

Next, we evaluate the impact of our proposed intermediate trajectory generation (Algorithm~\ref{alg:shielded-diffusion}, line~7) on task execution speed.
We conducted this ablation study without a person in the workspace, as the safety filter could otherwise slow down the robot.
To this end, we defined a modified version of the \textsc{Sorting} task, in which the green blocks are not picked up by a person.
Moreover, to eliminate the effect of the initial block placement on execution time, we always used the same starting configuration.
For comparison, we used the standard DP setup in which the generated actions are directly sent to the low-level controller at a frequency of \SI{30}{\hertz} instead of performing a time parameterization via~\eqref{eq:shield_intended_trajectory}.
The results of this experiment are provided in~\autoref{tab:sorting_results}.
Both approaches achieved the same success rate of~93\%, which is higher than in~\autoref{tab:safety_comparison}; likely due to the absence of the human and the fixed initial block configuration.
Yet, the intermediate trajectory generation reduced the average task execution time by~\SI{14}{\percent} from~\SI{25.2}{\second} to~\SI{21.7}{\second} and increased the average Cartesian end-effector speed by~\SI{13}{\percent}.
This \textbf{confirms~H4} and even shows that ensuring the kinematic and dynamic feasibility of the generated plan can lead to faster task execution without sacrificing performance.
{
\begin{table}[tb!]
    \centering
    \caption{Baseline comparison and speed analysis for the Sorting task.}
    \begin{tabular}{@{}llcc@{}}
        \toprule 
        Human & Safety Filter & Success & Duration $[\si{\second}]$ \\ 
        \midrule
        \multirow{2}{*}{w/ human} & Control barrier function & 0.43 & 35.8 $\pm$ 18.8 \\
        & PACS (\textbf{ours}) & \textbf{0.80} & \textbf{32.6 $\pm$ 11.1} \\
        \midrule
        \multirow{2}{*}{w/o human} & - & \textbf{0.93} & 25.2 $\pm$ 3.3 \\
        & PACS (\textbf{ours}) & \textbf{0.93} & \textbf{21.7 $\pm$ 3.9} \\
        \bottomrule
    \end{tabular}
    \label{tab:sorting_results}
\end{table}}

Lastly, we tested our safety filter for the \textsc{Feeding} task with a real human. As shown in our accompanying video, the robot slowed down when approaching the human, and the safeguarded policy inserted and withdrew the fork carefully and precisely.
These experiments highlight the potential of our approach to harness the expressiveness of DPs for highly safety-critical applications, such as healthcare.

\section{Conclusion}
We propose PACS, a general framework for safely deploying imitation learning policies using action chunks, such as DPs and vision-language-action models, in dynamic environments. 
Our insight is that safety interventions should be path-consistent to avoid out-of-distribution situations and maintain high task performance. 
We achieve this by computing an intended trajectory from the action chunk and monitoring it using reachability analysis.
Our experiments show that PACS can guarantee safety in real time and achieve high task success even on challenging real-world HRI tasks, whereas reactive safety filtering with control barrier functions frequently leads the policy into unrecoverable states.
Our method is specifically designed to satisfy dynamic constraints, such as moving obstacles. Handling (semi-)static obstacles via constraint-aware online replanning is an interesting avenue for future work.
We consider this work to be an important step towards deploying generative policies in safety-critical, human-centered environments.

\section*{Acknowledgements}
The authors gratefully acknowledge financial support of the research training group ConVeY funded by the German Research Foundation under grant GRK 2428; financial support by the DAAD program Konrad Zuse Schools of Excellence in Artificial Intelligence, sponsored by the Federal Ministry of Research, Technology and Space (BMFTR); and support by the Robotics Institute Germany, funded by BMFTR grant 16ME0997K.






\bibliographystyle{IEEEtran} 
\bibliography{library_cleaned} 

\end{document}